\newif\ifuseieee

\newcommand{\theabstract}{Accurate hand and finger tracking from video has significant clinical applications for monitoring activities of daily living and measuring range of motion, yet monocular video approaches for obtaining hand biomechanics remain under-developed. We present a method that combines the SAM 3D Body foundation model with inverse kinematics optimization in a full-body biomechanical model to extract anatomically-constrained finger joint angles from single-view video. We port SAM 3D Body from PyTorch to JAX for integration with MuJoCo-MJX, enabling GPU-accelerated optimization, and develop a novel mapping between the Momentum Human Rig (MHR) outputs and biomechanical model markers. Validation against 8-camera multiview reconstruction on 4,590 frames from 7 participants performing a variety of hand poses and object manipulation tasks shows finger joint angle errors of approximately 10 degrees and hand position errors of approximately 6 mm, after Procrustes alignment. Results were consistent across camera viewpoints and robust to different methods for producing reference values from multiview video. This work extends monocular biomechanical analysis to detailed finger tracking, expanding access to quantitative characterization of hand movement from readily available video.}

\ifuseieee

\documentclass[conference]{IEEEtran}
\IEEEoverridecommandlockouts

\usepackage{fontspec}
\usepackage{graphicx}
\usepackage{amsmath}
\usepackage{tabularx}
\usepackage{amsfonts}
\usepackage{amssymb}
\usepackage{gensymb}
\usepackage{booktabs}
\RequirePackage[colorlinks=true, allcolors=blue]{hyperref}
\usepackage{orcidlink}  

\usepackage[
    backend=biber,
    style=ieee,
    natbib=true,
    doi=false,
    url=false,
    isbn=false,
]{biblatex}

\AtEveryBibitem{%
   \clearfield{Title}%
   \clearfield{eprint}%
   \clearfield{note}%
}

\usepackage{etoolbox}

\makeatletter
\patchcmd{\@maketitle}{\Huge}{\LARGE}{}{}
\makeatother

\AtBeginEnvironment{tabular}{\scriptsize}
\renewcommand{\citet}[1]{\textcite{#1}}
\renewcommand{\citep}[1]{\parencite{#1}}

\else

\documentclass{isrpaper}

\usepackage{fontspec}

\usepackage{orcidlink}

\usepackage[
    backend=biber,
    style=authoryear,
    natbib=true,
    doi=false,
    url=false,
    isbn=false,
    maxbibnames=99,
    maxcitenames=2,
    uniquelist=false,
]{biblatex}

\AtEveryBibitem{%
   \clearfield{Title}%
   \clearfield{eprint}%
   \clearfield{note}%
}

\renewcommand{\citet}[1]{\textcite{#1}}
\renewcommand{\citep}[1]{\parencite{#1}}

\fi

\title{Monocular Biomechanical Tracking of Fingers with Inverse Kinematics to Foundation Models}

\ifuseieee

\author{
    \IEEEauthorblockN{1\textsuperscript{st        }R. James Cotton \orcidlink{0000-0001-5714-1400} }
    \IEEEauthorblockA{
            \textit{Shirley Ryan AbilityLab}\\       \textit{Department of PM\&R}\\       \textit{Northwestern University}            \\
      rcotton@sralab.org
          } \and     \IEEEauthorblockN{2\textsuperscript{nd        }Pouyan Firouzabadi}
    \IEEEauthorblockA{
            \textit{Shirley Ryan AbilityLab}\\       \textit{Department of Biomedical Engineering}\\       \textit{Northwestern University}            \\
      pfirouzaba@sralab.org
          } \and     \IEEEauthorblockN{3\textsuperscript{rd        }Wendy Murray}
    \IEEEauthorblockA{
            \textit{Shirley Ryan AbilityLab}\\       \textit{Department of Biomedical Engineering}\\       \textit{Northwestern University}            \\
      wmurray@sralab.org
          }}

\else

\author{
R. James Cotton\,\orcidlink{0000-0001-5714-1400}\,$^{1,2}$, Pouyan Firouzabadi$^{1,3}$, Wendy Murray$^{1,3}$}

\affiliations{
$^{1}$Shirley Ryan AbilityLab \\ $^{2}$Department of Physical Medicine and Rehabilitation, Northwestern University \\ $^{3}$Department of Biomedical Engineering, Northwestern University}

\paperdate{May 2026}

\correspondence{rcotton@sralab.org}

\paperabstract{\theabstract}

\fi

\begin{document}
\maketitle

\ifuseieee
\begin{abstract}\theabstract\end{abstract}
\fi


\section{Introduction}

Markerless motion capture (MMC) is advancing rapidly in many different dimensions. In addition to many studies showing that multiview MMMC shows a close agreement with marker-based methods in gait analysis it also shows a strong agreement with the shoulder and elbow \parencite{unger_differentiable_2025} and can even track finger joints within a few millimeters \parencite{firouzabadi_biomechanical_2024}. In parallel, monocular MMC is also improving rapidly, particularly for gait analysis and proximal joints in the upper extremity \parencite{peiffer_portable_2025, koleini_biopose_2025, xia_reconstructing_2025}. While there has also been remarkable progress in human pose estimation for tracking hands with mesh representations, this progress has not been translated into monocular approaches for obtaining hand biomechanics. The purpose of this study was to test whether biomechanical fits can be obtained from state-of-the-art models for human pose estimation, specifically the recently release Sam3d Body model \parencite{yang_sam_2025}, and to measure their accuracy for hand biomechanics.

Accurate hand and wrist tracking has significant clinical applications. It could enable continuous monitoring of activities of daily living (ADLs), provides objective measures of range of motion for conditions and allow detailed analysis of grasp patterns during functional tasks. This has the potential of providing a powerful biomarker for many far-ranging conditions.

Parametric human body models, such as SMPL \parencite{loper_smpl_2015}, models have played a massive role in human pose estimation. A key limitation of SMPL and its derivatives is that skeletal joint locations are derived from surface vertices, entangling the skeleton and body shape in ways that limit direct control over skeletal attributes. The Momentum Human Rig (MHR) \parencite{ferguson_mhr_2025} addresses this by explicitly decoupling the external body shape from the internal skeleton. MHR provides a kinematic tree with 127 joints controlled by 204 parameters (136 pose and 68 skeleton transformation), mesh vertices at multiple resolutions (up to 18,439 for the standard level of detail), and learned regressors that predict 308 keypoint locations corresponding to the Goliath set from Sapiens \parencite{khirodkar_sapiens_2024}. SAM 3D Body \parencite{yang_sam_2025} is a state-of-the-art inference model trained to predict MHR parameters from single images. It employs a dual-decoder architecture with a shared image encoder and separate decoders for body and hands, where a dedicated hand decoder processes high-resolution hand crops to achieve accurate finger pose estimation.

While SMPL and MHR are parametric body models designed to be compatible with computer vision research, there is an important parallel line of research on biomechanical models for human movement. The upper extremity model developed by Holzbaur and colleagues \parencite{holzbaur_model_2005} established a comprehensive musculoskeletal representation of the shoulder, elbow, forearm, and wrist. This was subsequently extended by \parencite{mcfarland_musculoskeletal_2023} to include all five digits of the hand with 43 Hill-type muscle-tendon actuators, creating the first open-source model capable of simulating functional tasks such as grip and pinch force production. There are also numerous biomechanical models of the lower extremities, such as the running model of Hamner et al. \parencite{hamner_muscle_2010} This has been ported to MuJoCo, a high-performance GPU accelerate framework in the LocoMuJoCo model \parencite{al-hafez_locomujoco_2023}. In this work, we combined a bilateral version of the upper extremity and hand model with the LocoMuJoCo lower body to produce a full-body biomechanical model with articulated hands, enabling inverse kinematics optimization from head to fingertips.

%

However, there are not powerful models such as SAM 3D Body that can directly predict accurate hand joint angles of a biomechanical model from images. Additionally, the accuracy of these models in terms of joint angles have not been characterized. Furthermore, there is a not a dictionary directly mapping keypoints and vertices from the MHR model onto the marker locations on our biomechanical model. In this work we developed such a mapping, used it to compute inverse kinematics for biomechanical models from monocular SAM 3D Body outputs, and compared these results against multiple multiview reconstructions.

\section{Methods}

\textbf{Overview:} Our pipeline for obtaining full-body biomechanics that involves the hand first passes images through the SAM 3D Body algorithms to regress the Momentum Human Rig (MHR) body configuration, followed by a multistage coarse-to-fine optimization that finds the inverse kinematics (IK) of that body configuration using a biomechanical model. We then compare the monocular fits against several multiview reference standards to estimate the errors from each individual view. We now elaborate on each component.

\subsection{JAX Implementation of SAM 3D Body and MHR}

Our recent biomechanical modeling has benefited from the massive parallelism and GPU acceleration of the MuJoCo-MJX framework \parencite{todorov_mujoco_2012}, which runs on Jax \parencite{jax_2018}. However, SAM 3D Body is implemented in pytorch \parencite{yang_sam_2025}. This would require our pipeline to require twice the GPU memory to allocate space for both Jax and Pytorch. To avoid this, we first ported the SAM 3D body architecture and model weights to Jax using the Equinox framework \parencite{kidger_equinox_2021}, allowing our entire pipeline to run in a single machine learning framework.

We refer to both the SAM 3D Body and MHR references for most details of these models \parencite{yang_sam_2025, ferguson_mhr_2025}, but elaborate on a few relevant details here. The MHR rig builds upon the ATLAS model \parencite{park_atlas_2025} and shares an important architectural difference from the widely used SMPL model \parencite{bogo_keep_2016} that is pertinent to biomechanics. Specifically, while SMPL model builds the joint center locations from scaled mesh before articulating it, MHR decouples scaling the kinematic tree (i.e., pseudo-skeleton) from the pose. This more closely follows the scaling and then posing approach used in biomechanics.

MHR includes 127 joints along the kinematic tree controlled by 136 pose parameters (rotational and translational) and 28 PCA skeleton scaling coefficients that span a 68-dimensional scaling space. While the underlying skeleton has more degrees of freedom than the human body at each joint (specifically 7 which includes 3 angular DOF, scaling, and also translational offsets), the 204 parameters reflect a sparse subset that are more physiological including interpretable bone length scaling and joint angle rotations, which also contrasts with SMPL. After posing the skeleton, the surface meshes vertex locations are computed with the addition of additional body scaling parameters. The surface mesh supports different levels of detail and we selected the model with 18,439 vertices. From these surface vertices, additional regressors produce 308 keypoint locations in 3D, which correspond to the Goliath keypoint set from Sapiens \parencite{khirodkar_sapiens_2024}; the first 70 of these (body, feet, and hand keypoints) form the subset used downstream in this work. Regressors allow predicting the location of keypoints that are not on the kinematic tree and also may not be on the surface of the body, such as the various definitions for ``hip'' used in computer vision.


SAM 3D Body is a model trained on a massive amount of data to predict the MHR parameters from images \parencite{yang_sam_2025}. It was trained on two backbones that showed similar performance, and we ported the model variant based on the DINOv3 backbone \parencite{simeoni_dinov3_2025}. It supports several optional inputs for refining the predictions, including masks or 2D keypoints. We did not explore using these additional components in this work. It is also a `top-down' model and expects to receive a cropped and centers image of a person. We used the bounding boxes from PosePipe within our multicamera framework to provide these cropped and centered images \parencite{cotton_posepipe_2022, cotton_differentiable_2025}.  SAM 3D also has a hand specific encoder and can be used in a coarse-to-fine manner using first a full body image to provide a first-pass model of the pose and then to zoom into the hands for a refined MHR estimate of the hand, which is merged to provide a refined estimate. Our Equinox port of SAM 3D Body supported this recursive inference and hand refinement, which was used in this work.

\subsection{Full body biomechanical model with hand markers}

Much of our prior work has focused on gross body movements without finger articulation \parencite{peiffer_portable_2025, cotton_differentiable_2025}, using the LocoMuJoCo model which is built upon the Hamner OpenSim model \parencite{al-hafez_locomujoco_2023, hamner_muscle_2010, delp_opensim_2007}. One exception was our work showing that we can reconstruct finger movements within a few millimeters from our multiview camera acquisition system \parencite{firouzabadi_biomechanical_2024}. For this, we used the MOBL-ARMS model, which is the first neuromuscular biomechanical model of the upper extremity capable of simulating functional tasks \parencite{mcfarland_musculoskeletal_2023}. The MOBL-ARMS model includes the rib cage and right arm. In this work, we used a combined model which includes a bilateral fully articulated arm and hand model with a LocoMuJoCo legs model, totaling 111 generalized coordinates (nq) across 83 bodies and 105 joints. This allowed us to align the full body and hand MHR output from SAM 3D Body with a full body and hand biomechanical model.

One critical barrier was the lack of a mapping between these two different models for optimizing their similarity. In our prior work, we optimized the position of 87 markers on the LocoMuJoCo model that correspond to the MoVI keypoints \parencite{ghorbani_movi_2021, cotton_differentiable_2025}, in order to align our biomechanical models with state of the art keypoint detection algorithms, MeTRAbs \parencite{sarandi_learning_2023}. In this work we took the opposite approach: find the set of outputs from SAM 3D Body that corresponded to anatomical landmark sites defined on our biomechanical model. Specifically, we took videos of people walking from our database that were processed using both MeTRAbs and SAM 3D and found a mapping of either MHR mesh vertices or regressed keypoints that most closely matched the MoVI keypoint positions. This produced 73 body marker correspondences, where each MHR output (mesh vertex or predicted keypoint) is paired with a site on the biomechanical skeleton. The mapping for the hands was more straightforward as there is a direct anatomical correspondence between finger joint centers in both model architectures. The MCP, PIP, and DIP joints are semantically equivalent between MHR and the biomechanical model, so these markers are placed at (or within a few millimeters of) the corresponding joint centers on both sides. The fingertips are placed as site markers on the distal phalanges beyond the last joint --- these are not on the kinematic tree but provide critical constraints for finger pose. One additional marker per hand at the wrist completes the set, yielding 42 hand markers total (21 per hand: wrist, thumb CMC and IP, finger MCP/PIP/DIP joint centers, and 5 fingertips). All 42 hand markers are matched to MHR's regressed hand keypoints rather than mesh vertices as these sites are internal to the mesh surface. The same anatomical principle holds for the body markers: the knee markers (LKnee, RKnee) are placed essentially at the tibia joint center and, along with LAnkle and Neck, are matched to MHR's regressed body keypoints rather than mesh vertices. The 5 face landmarks (nose, eyes, ears) are likewise drawn from MHR's regressed face keypoints. In total the full mapping contains 120 correspondences between MHR outputs and biomechanical model sites: 69 mesh-vertex correspondences and 51 regressed-keypoint correspondences (42 hand + 5 face + 4 body joint centers). The body marker site positions on the skeleton were further refined using an EM-optimization procedure, and the residual error between the two systems was used to determine the confidence weight applied to each marker.


\subsection{Levenberg-Marquardt IK Optimization}

The mapping between the full body MuJoCo model and the MHR model allowed us to measure a loss function between the IK fit of the MuJoCo model and the MHR target pose. This loss was:

\begin{equation}
\mathcal{L} = \sum_{i} w_i \| \mathbf{p}_i^{\text{pred}} - \mathbf{p}_i^{\text{model}}(\mathbf{q}) \|^2 + \lambda_{\text{reg}} \| \boldsymbol{\delta} \|^2
\end{equation}

Where $\mathbf{q}$ represents the joint angles, $\mathbf{p}_i^{\text{pred}}$ are the Sam3D vertex and regressed keypoint locations, $\mathbf{p}_i^{\text{model}}(\mathbf{q})$ are the marker positions computed via forward kinematics, and $\boldsymbol{\delta}$ are per-marker offsets regularized by $\lambda_{\text{reg}}$ to prevent overfitting. Joint angles are also subject to a soft penalty for ball-joint limit violations in addition to hard clamping to anatomical ranges after each LM step.

We minimized this loss using the Levenberg-Marquardt algorithm with a multistage coarse-to-fine optimization. The solver was warm-started from MHR's predicted body root pose, with the root translation then replaced by the centroid of the projected MHR marker targets. Stage 1 optimizes only the pelvis position and orientation using a subset of core body markers (pelvis, spine, shoulder, upper arm, elbow, and forearm landmarks). Stage 2 optimizes all joint angles including finger DOFs with body segment scales held fixed. Stage 3 jointly optimizes pose, body segment scales, and per-marker offsets, where the offsets compensate for differences in marker placement between the MHR and biomechanical models. JAX automatically differentiates through the MuJoCo forward kinematic model to efficiently solve this second order optimization problem.

The monocular IK solver used a damping factor of $\lambda = 3.0$, with 400 total iterations distributed evenly across the three active stages. Per-marker offset regularization used $\lambda_{\text{reg}} = 20.0$.

\subsection{Multiview Reconstruction for Reference Poses}

For our reference values we used three versions. One was previously described \parencite{firouzabadi_biomechanical_2024}. Briefly, it is an end to end algorithm that directly optimizes for the pose that minimizes the reprojection errors. It includes an initial detection of the whole body followed by a specialized hand detector for the hand keypoints. This optimization involves scaling the skeleton jointly while also fitting the kinematics through time over all of the trials. This method was adopted to maintain a consistent scaling of the musculoskeletal model representing the participant for a more biomechanically accurate analysis.

We also compared to two other methods for multiview reconstruction. This was to compare more modern keypoint detectors to see if they provide stronger reference values and ensure our results are robust to the reference values. Obtaining true ground truth with biplanar fluorescopy is very challenging. For these we extracted the SAM 3D Body representation from each view and then used each of those keypoints in the 2D image plane reference frame. We then ran LM optimization for the pose with the objective of minimizing the reprojection error across cameras:

\begin{equation}
\mathbf{q}^* = \arg\min_{\mathbf{q}} \sum_{c=1}^{C} \sum_{k=1}^{K} w_{c,k} \left\| \pi_c\bigl(\text{FK}(\mathbf{q})_k\bigr) - \mathbf{y}_{c,k} \right\|^2
\end{equation}

where $\pi_c$ is the projection function for camera $c$, $\text{FK}(\mathbf{q})_k$ is the 3D position of marker $k$ from forward kinematics, $\mathbf{y}_{c,k}$ is the detected 2D keypoint, and $w_{c,k}$ is the per-keypoint-per-camera weight. The multiview optimizer used damping $\lambda = 1.0$ with 500 LM iterations across three stages (30\% root positioning, 50\% full pose, 20\% pose with scale optimization), followed by a final 20-iteration stage that refined per-frame marker offsets only with $\lambda_{\text{reg}} = 10^4$ (pose and scale held fixed). This was computed both using the markers from Sam3D Body and those from the Sapiens model \parencite{khirodkar_sapiens_2024}.

\textbf{Occlusion Handling} we used several mechanisms to make the multiview optimization robust to occlusion and outliers. First, each marker carries a static confidence weight derived from the mapping procedure, and any keypoints with confidence below 0.25 or outside the image frame are zeroed. Second, we use robust triangulation across the views for each individual keypoint to measure the geometric consensus across cameras; cameras that are inconsistent for a given keypoint are downweighted via a Gaussian kernel at a per-keypoint-per-camera granularity \parencite{cotton_improved_2023}. This implicitly handles self-occlusion, since a keypoint that is occluded in one view produces an inconsistent 2D detection that is automatically downweighted by the triangulation consensus.

In all cases we computed the $GC_{10}$, which is the number of views where the reconstructed 3D locations on the skeleton from the marker reproject within 10 pixels on the original image as a metric to indicate high quality fits \parencite{cotton_improved_2023}.

\subsection{Evaluation Metrics}

We compare monocular reconstructions against the multiview reference using Procrustes-aligned mean per-joint position error (PA-MPJPE), which removes global rotation, translation, and scale differences before computing 3D Euclidean distances. For the upper extremity, we jointly align all upper-extremity sites (shoulder, upper arm, elbow, forearm, wrist, and hand/finger markers). For hands, we perform a separate Procrustes alignment on hand markers only, isolating finger pose accuracy from arm positioning errors. We also report angular errors as the mean absolute difference in joint angles.

The primary comparison table reports: (1) upper extremity PA-MPJPE in millimeters, (2) hand-aligned PA-MPJPE averaged across left and right hands, and (3) geometric consistency (GC@10px)---the percentage of keypoints whose multiview reprojection error falls below 10 pixels, which indicates the reliability of the multiview reference.

Supplementary analyses include per-finger breakdowns of angular and positional error, upper-extremity-aligned hand error (which captures combined arm and finger error), and cross-view consistency measuring the standard deviation of monocular joint angle estimates across camera views.

\subsection{Evaluation Dataset}

We evaluate on a dataset from a multiview hand motion capture collection designed for fine-grained finger tracking validation \parencite{firouzabadi_biomechanical_2024}:

\begin{itemize}
\item \textbf{Participants}: 7 individuals
\item \textbf{Recordings}: 221 multiview sessions
\item \textbf{Cameras}: 8 synchronized and calibrated views per session
\item \textbf{Activities}: American Sign Language (ASL) finger spelling, number counting gestures, isolated finger movements (abduction, PIP/MCP flexion, wrist movements, lumbricals), fine motor tasks (key manipulation, pin relocation, writing, sphere and disk grasping), and naturalistic object manipulation (bottle grasping, box handling, drinking)
\item \textbf{Sampled frames}: 4,590 video frames (stride of 30 frames)
\item \textbf{Total camera-frames}: 36,720 (video frames $\times$ 8 cameras)
\end{itemize}

The dataset captures diverse hand poses across participants with varying hand sizes and movement styles. Each session provides multiview coverage enabling robust 3D reference reconstruction through triangulation and reprojection optimization.

\section{Results}

\subsection{Example Reconstructions}

\begin{figure}[!htbp]
\centering
\includegraphics[width=1\linewidth]{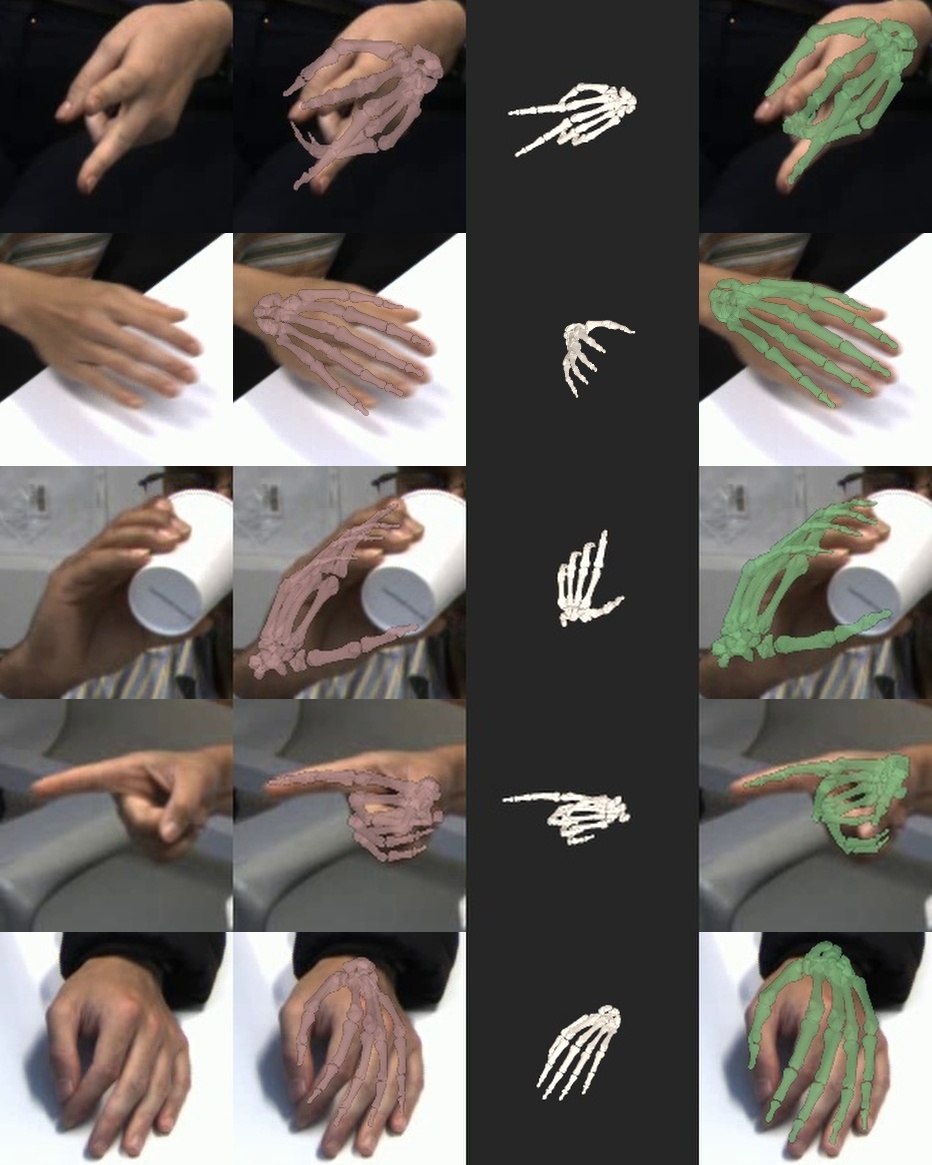}
\caption[]{Hand reconstruction examples across participants. Each row shows one participant with panels for: raw image, monocular overlay (red), top-down view, and multiview reference overlay (green).}
\label{fig-hand-closeup}
\end{figure}

Figure~\ref{fig-hand-closeup} shows representative hand reconstructions comparing monocular and multiview results. Each row displays a different participant performing various hand gestures. The four panes show: (1) the raw cropped hand image, (2) the monocular Sam3D + IK reconstruction overlaid in red, (3) a skeleton rendering, and (4) the multiview reference reconstruction overlaid in green. The monocular reconstructions capture the overall hand configuration, though subtle differences in finger flexion angles are visible when compared to the geometrically-constrained multiview fits.

\subsection{Reference Target Comparison}

We compared monocular predictions against three different multiview references to assess the accuracy of our measurements and ensure it is consistent across methods: 1) End-to-end Reconstruction using \parencite{firouzabadi_biomechanical_2024} applied to the entire time series.
2) Sam3D \parencite{yang_sam_2025} multiview by directly optimizing the biomechanical model using using body and hand keypoints in the 2D image plane, which was performed only on selected frames independently 3) Sapiens (1B model) multiview by using the hand keypoints from \parencite{khirodkar_sapiens_2024} but otherwise the same as 2.

\begin{table}
\centering
\caption[]{Comparison of monocular accuracy against different multiview reference methods.

\begin{tabular}{p{\dimexpr 0.200\linewidth-2\tabcolsep}p{\dimexpr 0.200\linewidth-2\tabcolsep}p{\dimexpr 0.200\linewidth-2\tabcolsep}p{\dimexpr 0.200\linewidth-2\tabcolsep}p{\dimexpr 0.200\linewidth-2\tabcolsep}}
\toprule
Reference & UE PA (mm) & Hand PA (mm) & Hand Angle (°) & GC@10px \\
\hline
End-to-end & 23.8 $\pm$ 13.3 & 8.5 $\pm$ 4.3 & 11.4 $\pm$ 5.6 & 47.6\% \\
Sam3D MV & 14.3 $\pm$ 6.4 & 4.8 $\pm$ 2.4 & 7.8 $\pm$ 3.2 & 55.6\% \\
Sapiens MV & 15.8 $\pm$ 7.7 & 5.9 $\pm$ 3.0 & 8.3 $\pm$ 3.9 & 56.0\% \\
\bottomrule
\end{tabular}}
\label{tbl-3method-comparison}
\end{table}

In general, the results from the IK fits to the monocular MHR results (Table~\ref{tbl-3method-comparison}) show fairly similar values across the different reference targets, which supports the robustness of the monocular results. The geometric consistency at 10 pixels ($GC_{10}$) of the multiview reconstructions was slightly greater than our prior approach \parencite{firouzabadi_biomechanical_2024}. This could the slightly more challenging constraints of the end-to-end optimization, which requires a single skeleton scale over consistently for each frame over all videos from a participant. It likely also reflects some improvements in human pose estimation algorithms in the last year.

Using Sapiens as the multiview reference value, we see that the mean errors throughout the arm and hand are below 25 mm and when Procrustes alignment is applied specifically to the hand are about 6mm. The joint angles throughtout the hand show an error of about 8 degrees.

The similarity of the results between Sam3D and Sapiens is also reassuring, as keypoint detection tends to track fine details in the image more precisely than mesh regression. The Sapiens multiview reference also has the methodological advantage of being more independent of the monocular IK pipeline, which is driven by Sam3D outputs. The two references agree closely (within {\textasciitilde}1.5 mm in UE PA-MPJPE), so any Sam3D-side systematic bias is small in practice, but for the remainder of the analysis we use the Sapiens multiview as the primary reference.

\subsection{Detailed Analysis (Sapiens Multiview Reference Value)}

The following analysis uses Sapiens multiview reconstruction as reference truth. The multiview system provides 3D keypoint locations and joint angles by triangulating observations across 8 synchronized camera views.

\subsubsection{Overall Accuracy}

\begin{figure}[!htbp]
\centering
\includegraphics[width=1\linewidth]{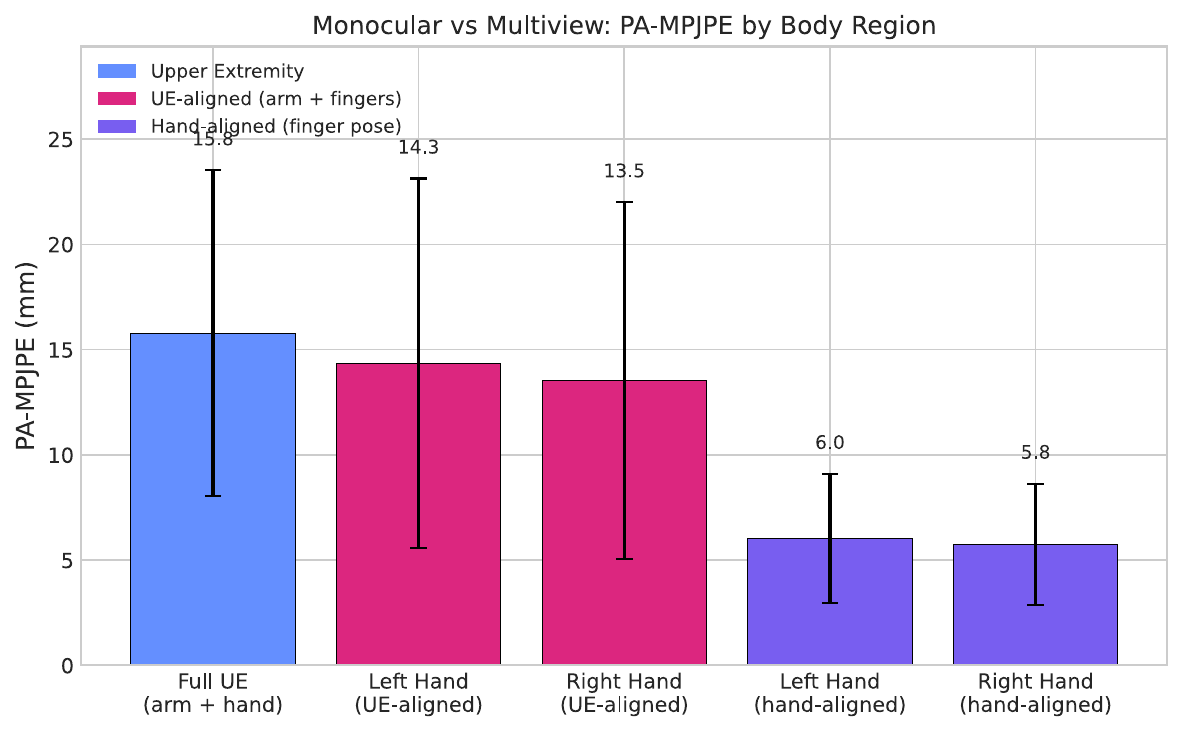}
\caption[]{Comparison of PA-MPJPE across body regions. The Full UE bar averages over all upper-extremity markers (arm + hand). Hand-only bars subset this: UE-aligned keeps the upper-extremity Procrustes alignment but reports hand markers only ({\textasciitilde}14 mm); hand-aligned uses a hand-only alignment to isolate finger articulation accuracy ({\textasciitilde}6 mm). Hand markers fit the alignment more tightly than the proximal arm/torso markers, so the Full UE total exceeds the UE-aligned hand error.}
\label{fig-accuracy-comparison}
\end{figure}

We evaluated the monocular tracking accuracy using Procrustes-Aligned Mean Per-Joint Position Error (PA-MPJPE), which measures 3D joint position accuracy after optimal alignment to remove global position and orientation differences. Figure~\ref{fig-accuracy-comparison} shows the accuracy across body regions. The \textbf{Full UE} bar averages over all upper-extremity markers --- both the proximal arm/torso markers (shoulder, scapula, elbow, forearm, wrist) and the 42 hand markers --- under a single Procrustes alignment of the full set. The hand-only bars subset this measurement: \textbf{UE-aligned} keeps the same upper-extremity alignment but reports the error on hand markers only, while \textbf{hand-aligned} performs a separate Procrustes alignment using only the hand markers (isolating finger articulation accuracy from arm-positioning error).

\subsubsection{Per-Joint Analysis}

\begin{figure}[!htbp]
\centering
\includegraphics[width=1\linewidth]{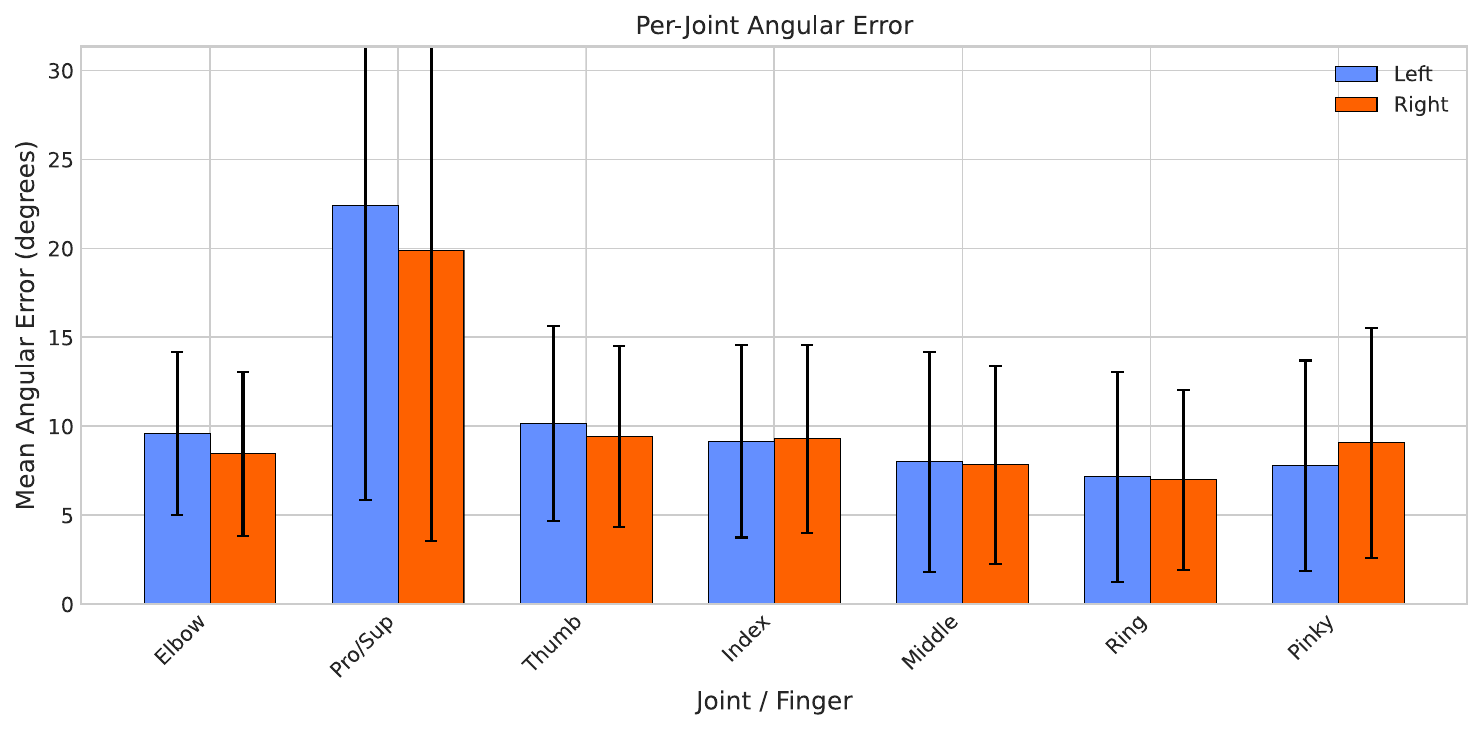}
\caption[]{Per-joint mean angular error for left and right upper extremities, computed as the absolute difference between monocular IK joint angles and the corresponding Sapiens multiview IK joint angles (Sapiens-substituted hand keypoints with the same biomechanical model fit across 8 views). Includes elbow flexion, forearm pronation/supination, and individual finger errors.}
\label{fig-per-finger-angular}
\end{figure}

Figure~\ref{fig-per-finger-angular} shows the angular errors broken down by joint for both left and right sides. The greatest error comes from the supination pronation joint with the hand joints all showing mean errors below 10 degrees.

\subsubsection{Per-Camera Analysis}

\begin{figure}[!htbp]
\centering
\includegraphics[width=1\linewidth]{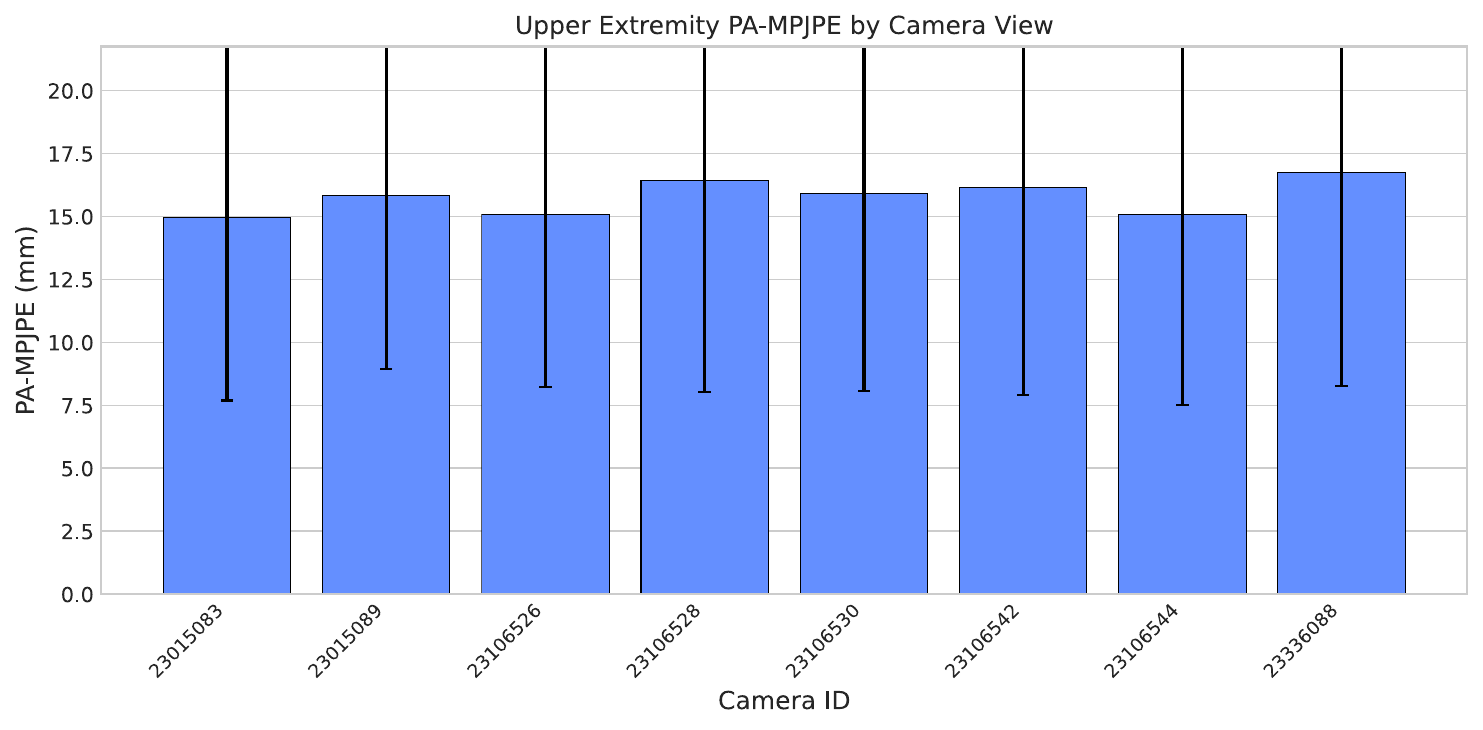}
\caption[]{Upper extremity PA-MPJPE by camera viewpoint.}
\label{fig-camera-comparison}
\end{figure}

Figure~\ref{fig-camera-comparison} shows the accuracy broken down by camera viewpoint. These cameras span a wide range of viewpoint and angles, although views and frames where the camera was occluded were excluded. The results show minimal evidence of viewpoint dependence.

\section{Discussion}

This work demonstrates how foundation models for human mesh reconstruction can be combined with biomechanical inverse kinematics to enable detailed finger tracking from monocular video. Our multistage Levenberg-Marquardt optimization approach (root positioning, full pose with scaling, and marker offset refinement) provides robust convergence and accurate joint angle estimation. Furthermore, porting SAM 3D Body to Equinox and JAX allows the entire pipeline to run efficiently on a single GPU. The auto-differentiation from MuJoCo-MJX also makes implementing LM optimization straightforward.

Our validation against multiview reconstruction shows that monocular predictions achieve reasonable accuracy compared to multiview reference data. We make an important distinction to call this reference data rather than ground truth as the markerless motion capture does have errors. However, practically obtaining marker-based data for each joint of the hand is both impractical and visually changes the appearance making such comparisons unlikely to reflect generalization to unmarked settings. The stability of the monocular fits against different multiview methods (end-to-end optimization to entire time series data versus individual frame multiview optimization with different keypoint sets) is further reassuring that this at least places a bound on the accuracy of the monocular results.

Whether the accuracy we describe is sufficient will depend greatly on the clinical question and application domain. For example, the errors on the order of 10 degrees in the fingers (with larger errors near 20 degrees in forearm pronation/supination) might be insufficient for tracking range of motion longitudinally through time. Conversely, it is also much more precise than could be done from monocular videos even several years ago. It is also important to note that the standard deviations in our errors were fairly wide, and there were many trials where the accuracy was much worse than the average. An important future direction, in addition to improving the accuracy, is incorporating reliable quality signals that indicate when the results are trustworthy \parencite{cotton_biomechanical_2025, donahue_embc_2026}.

An important limitation of this work is that it does not contain any data from participants with motor impairments in hand function or visible limb differences. For the former, is it very possible that hand postures that might be common for people with motor impairments but are atypical in the training data of SAM 3D Body would perform differently. Additionally, it is unlikely that this level of performance would be obtained for people with missing digits or using prostheses.

Finally, more accessible biomechanical analysis of hand movements from monocular video opens up a whole range of novel problems. The range of things people do with their hands is enormous, and making sense of this data to guide care is a challenging problem. Furthermore, an important additional issue is understanding what objects people are interacting with in addition to the posture of the hand. While a few of our activities involved hands interacting with objects, such as the drinking task, we did not systematically evaluate the influence of this on hand tracking accuracy. An important future direction will be incorporating models that fit hand-object interactions from video with biomechanical modeling (e.g., \parencite{fan_hold_2023, wu_reconstructing_2024}).

\section{Conclusion}

We presented a method for monocular biomechanical finger tracking that combines Sam3D foundation model keypoints with inverse kinematics optimization in MuJoCo-MJX. The JAX implementation enables efficient GPU-accelerated processing, and validation against multiview reconstruction demonstrates accurate joint angle estimation. This approach extends monocular biomechanical analysis to detailed hand and finger tracking, with applications to clinical assessment and infant movement analysis.



\ifuseieee
\section*{Acknowledgements}
\footnotesize
\else
\section*{Acknowledgments}
\small
\fi

This work was supported by the Shirley Ryan AbilityLab research accelerator program.

\normalsize

{\small
\printbibliography

@article{firouzabadi_biomechanical_2024,
  title = {Biomechanical Arm and Hand Tracking with Multiview Markerless Motion Capture},
  author = {Firouzabadi, Pouyan and Murray, Wendy and Sobinov, Anton R and Peiffer, J.D. and Shah, Kunal and Miller, Lee E and Cotton, R. James},
  journal = {IEEE International Conference on Biomedical Robotics and Biomechatronics (BioRob)},
  year = {2024},
  doi = {10.1109/BioRob60516.2024.10719940},
  note = {Reference for multiview validation methodology}
}

@inproceedings{loper_smpl_2015,
  title = {{SMPL}: A Skinned Multi-Person Linear Model},
  author = {Loper, Matthew and Mahmood, Naureen and Romero, Javier and Pons-Moll, Gerard and Black, Michael J.},
  booktitle = {ACM Transactions on Graphics (Proc. SIGGRAPH Asia)},
  year = {2015},
  doi = {10.1145/2816795.2818013}
}

@inproceedings{todorov_mujoco_2012,
  title = {{MuJoCo}: A Physics Engine for Model-Based Control},
  author = {Todorov, Emanuel and Erez, Tom and Tassa, Yuval},
  booktitle = {IEEE/RSJ International Conference on Intelligent Robots and Systems},
  year = {2012},
  doi = {10.1109/IROS.2012.6386109}
}

@software{jax_2018,
  title = {{JAX}: Composable Transformations of {Python}+{NumPy} Programs},
  author = {Bradbury, James and Frostig, Roy and Hawkins, Peter and Johnson, Matthew James and Leary, Chris and Maclaurin, Dougal and Necula, George and Paszke, Adam and VanderPlas, Jake and Wanderman-Milne, Skye and Zhang, Qiao},
  year = {2018},
  url = {https://github.com/google/jax}
}

@article{hamner_muscle_2010,
  title = {Muscle contributions to propulsion and support during running},
  author = {Hamner, Samuel R. and Seth, Ajay and Delp, Scott L.},
  journal = {Journal of Biomechanics},
  volume = {43},
  number = {14},
  pages = {2709--2716},
  year = {2010},
  doi = {10.1016/j.jbiomech.2010.06.025}
}

@article{ghorbani_movi_2021,
  title = {{MoVi}: A Large Multi-Purpose Human Motion and Video Dataset},
  author = {Ghorbani, Saeed and Mahdaviani, Kimia and Thaler, Anne and Kording, Konrad and Cook, Douglas James and Blohm, Gunnar and Troje, Nikolaus F.},
  journal = {PLOS ONE},
  volume = {16},
  number = {6},
  pages = {e0253157},
  year = {2021},
  doi = {10.1371/journal.pone.0253157}
}

@article{unger_differentiable_2025,
  title = {Differentiable biomechanics for markerless motion capture in upper limb stroke rehabilitation: a comparison with optical motion capture},
  author = {Unger, Tim and Moslehian, Arash Sal and Peiffer, J.D. and Ullrich, Johann and Gassert, Roger and Lambercy, Olivier and Cotton, R. James and Easthope, Chris Awai},
  journal = {IEEE Transactions on Medical Robotics and Bionics},
  year = {2025},
  doi = {10.1109/TMRB.2025.3605962},
  pages = {1--1}
}

@misc{koleini_biopose_2025,
  title = {{BioPose}: {Biomechanically}-accurate {3D} {Pose} {Estimation} from {Monocular} {Videos}},
  author = {Koleini, Farnoosh and Saleem, Muhammad Usama and Wang, Pu and Xue, Hongfei and Helmy, Ahmed and Fenwick, Abbey},
  year = {2025},
  doi = {10.48550/arXiv.2501.07800},
  url = {http://arxiv.org/abs/2501.07800},
  note = {arXiv:2501.07800}
}

@misc{yang_sam_2025,
  title = {{SAM} {3D} {Body}: {Robust} {Full}-{Body} {Human} {Mesh} {Recovery}},
  author = {Yang, Xitong and Kukreja, Devansh and Pinkus, Don and Sagar, Anushka and Fan, Taosha and Park, Jinhyung and Cao, Jinkun and Liu, Jiawei and Ugrinovic, Nicolas and Feiszli, Matt and Malik, Jitendra and Dollar, Piotr and Kitani, Kris},
  year = {2025},
  doi = {10.48550/arXiv.2511.15586},
  url = {http://arxiv.org/abs/2511.15586},
  note = {arXiv:2511.15586}
}

@misc{ferguson_mhr_2025,
  title = {{MHR}: {Momentum} {Human} {Rig}},
  author = {Ferguson, Aaron and Osman, Ahmed A. A. and Bescos, Berta and Stoll, Carsten and Twigg, Chris and Lassner, Christoph and Otte, David and Vignola, Eric and Prada, Fabian and Bogo, Federica and Santesteban, Igor and Romero, Javier and Zarate, Jenna and Lee, Jeongseok and Park, Jinhyung and Yang, Jinlong and Doublestein, John and Venkateshan, Kishore and Kitani, Kris and Kavan, Ladislav and Farra, Marco Dal and Hu, Matthew and Cioffi, Matthew and Fabris, Michael and Ranieri, Michael and Modarres, Mohammad and Kadlecek, Petr and Khirodkar, Rawal and Abdrashitov, Rinat and Prevost, Romain and Rajbhandari, Roman and Mallet, Ronald and Pearsall, Russel and Kao, Sandy and Kumar, Sanjeev and Parrish, Scott and Yu, Shoou-I. and Saito, Shunsuke and Shiratori, Takaaki and Wang, Te-Li and Tung, Tony and Xu, Yichen and Dong, Yuan and Chen, Yuhua and Xu, Yuanlu and Ye, Yuting and Jiang, Zhongshi},
  year = {2025},
  doi = {10.48550/arXiv.2511.15586},
  url = {http://arxiv.org/abs/2511.15586},
  note = {arXiv:2511.15586}
}

@article{holzbaur_model_2005,
  title = {A model of the upper extremity for simulating musculoskeletal surgery and analyzing neuromuscular control},
  author = {Holzbaur, Katherine R. S. and Murray, Wendy M. and Delp, Scott L.},
  journal = {Annals of Biomedical Engineering},
  volume = {33},
  number = {6},
  pages = {829--840},
  year = {2005},
  doi = {10.1007/s10439-005-3320-7}
}

@article{mcfarland_musculoskeletal_2023,
  title = {A {Musculoskeletal} {Model} of the {Hand} and {Wrist} {Capable} of {Simulating} {Functional} {Tasks}},
  author = {McFarland, Daniel C. and Binder-Markey, Benjamin I. and Nichols, Jennifer A. and Wohlman, Sarah J. and de Bruin, Marije and Murray, Wendy M.},
  journal = {IEEE transactions on bio-medical engineering},
  volume = {70},
  number = {5},
  pages = {1424--1435},
  year = {2023},
  doi = {10.1109/TBME.2022.3217722}
}

@inproceedings{cotton_differentiable_2025,
  title = {Differentiable {Biomechanics} {Unlocks} {Opportunities} for {Markerless} {Motion} {Capture}},
  author = {Cotton, R. James},
  booktitle = {2025 International Conference On Rehabilitation Robotics (ICORR)},
  pages = {44--51},
  year = {2025},
  doi = {10.1109/ICORR66766.2025.11063174},
  url = {https://ieeexplore.ieee.org/document/11063174}
}

@incollection{bogo_keep_2016,
  title = {Keep {It} {SMPL}: {Automatic} {Estimation} of {3D} {Human} {Pose} and {Shape} from a {Single} {Image}},
  author = {Bogo, Federica and Kanazawa, Angjoo and Lassner, Christoph and Gehler, Peter and Romero, Javier and Black, Michael J.},
  booktitle = {ECCV 2016: Computer Vision -- ECCV 2016},
  publisher = {Springer, Cham},
  pages = {561--578},
  year = {2016},
  doi = {10.1007/978-3-319-46454-1_34}
}

@misc{peiffer_portable_2025,
  title = {Portable biomechanics laboratory: {Clinically} accessible movement analysis from a handheld smartphone},
  author = {Peiffer, J. D. and Shah, Kunal and Djuraskovic, Irina and Anarwala, Shawana and Abdou, Kayan and Patel, Rujvee and Jayabalan, Prakash and Pennicooke, Brenton and Cotton, R. James},
  year = {2025},
  url = {https://arxiv.org/abs/2507.08268},
  note = {arXiv:2507.08268}
}

@inproceedings{xia_reconstructing_2025,
  title = {Reconstructing {Humans} with a {Biomechanically} {Accurate} {Skeleton}},
  author = {Xia, Yan and Zhou, Xiaowei and Vouga, Etienne and Huang, Qixing and Pavlakos, Georgios},
  booktitle = {CVPR},
  year = {2025},
  doi = {10.48550/arXiv.2503.21751},
  url = {http://arxiv.org/abs/2503.21751},
  note = {arXiv:2503.21751}
}

@article{delp_opensim_2007,
  title = {{OpenSim}: Open-Source Software to Create and Analyze Dynamic Simulations of Movement},
  author = {Delp, Scott L. and Anderson, Frank C. and Arnold, Allison S. and Loan, Peter and Habib, Ayman and John, Chand T. and Guendelman, Eran and Thelen, Darryl G.},
  journal = {IEEE Transactions on Biomedical Engineering},
  volume = {54},
  number = {11},
  pages = {1940--1950},
  year = {2007},
  doi = {10.1109/TBME.2007.901024}
}

@misc{khirodkar_sapiens_2024,
  title = {Sapiens: Foundation for Human Vision Models},
  author = {Khirodkar, Rawal and Bagautdinov, Timur and Martinez, Julieta and Zhaoen, Su and James, Austin and Selednik, Peter and Anderson, Stuart and Saito, Shunsuke},
  year = {2024},
  doi = {10.48550/arXiv.2408.12569},
  url = {http://arxiv.org/abs/2408.12569},
  note = {arXiv:2408.12569}
}

@misc{park_atlas_2025,
  title = {{ATLAS}: Decoupling Skeletal and Shape Parameters for Expressive Parametric Human Modeling},
  author = {Park, Jinhyung and Romero, Javier and Saito, Shunsuke and Prada, Fabian and Shiratori, Takaaki and Xu, Yichen and Bogo, Federica and Yu, Shoou-I. and Kitani, Kris and Khirodkar, Rawal},
  year = {2025},
  doi = {10.48550/arXiv.2508.15767},
  url = {http://arxiv.org/abs/2508.15767},
  note = {arXiv:2508.15767}
}

@misc{simeoni_dinov3_2025,
  title = {{DINOv3}},
  author = {Siméoni, Oriane and Vo, Huy V. and Seitzer, Maximilian and Baldassarre, Federico and Oquab, Maxime and Jose, Cijo and Khalidov, Vasil and Szafraniec, Marc and Yi, Seungeun and Ramamonjisoa, Michaël and Massa, Francisco and Haziza, Daniel and Wehrstedt, Luca and Wang, Jianyuan and Darcet, Timothée and Moutakanni, Théo and Sentana, Leonel and Roberts, Claire and Vedaldi, Andrea and Tolan, Jamie and Brandt, John and Couprie, Camille and Mairal, Julien and Jégou, Hervé and Labatut, Patrick and Bojanowski, Piotr},
  year = {2025},
  doi = {10.48550/arXiv.2508.10104},
  url = {http://arxiv.org/abs/2508.10104},
  note = {arXiv:2508.10104}
}

@inproceedings{cotton_improved_2023,
  title = {Improved Trajectory Reconstruction for Markerless Pose Estimation},
  author = {Cotton, R. James and Cimorelli, Anthony and Shah, Kunal and Anarwala, Shawana and Uhlrich, Scott and Karakostas, Tasos},
  booktitle = {45th Annual International Conference of the IEEE Engineering in Medicine and Biology Society},
  year = {2023},
  url = {https://ieeexplore.ieee.org/document/10340745}
}

@article{kidger_equinox_2021,
  title = {Equinox: neural networks in {JAX} via callable {PyTrees} and filtered transformations},
  author = {Kidger, Patrick and Garcia, Cristian},
  journal = {Differentiable Programming workshop at Neural Information Processing Systems 2021},
  year = {2021},
  url = {https://arxiv.org/abs/2111.00254}
}

@misc{al-hafez_locomujoco_2023,
  title = {{LocoMuJoCo}: A Comprehensive Imitation Learning Benchmark for Locomotion},
  author = {Al-Hafez, Firas and Zhao, Guoping and Peters, Jan and Tateo, Davide},
  year = {2023},
  doi = {10.48550/arXiv.2311.02496},
  url = {http://arxiv.org/abs/2311.02496},
  note = {arXiv:2311.02496}
}

@inproceedings{sarandi_learning_2023,
  title = {Learning 3D Human Pose Estimation from Dozens of Datasets using a Geometry-Aware Autoencoder to Bridge Between Skeleton Formats},
  author = {Sárándi, István and Hermans, Alexander and Leibe, Bastian},
  booktitle = {IEEE/CVF Winter Conference on Applications of Computer Vision (WACV)},
  year = {2023},
  doi = {10.48550/arXiv.2212.14474},
  url = {http://arxiv.org/abs/2212.14474}
}

@article{cotton_posepipe_2022,
  title = {{PosePipe}: Open-Source Human Pose Estimation Pipeline for Clinical Research},
  author = {Cotton, R. James},
  journal = {arXiv preprint arXiv:2203.08792},
  year = {2022},
  url = {http://arxiv.org/abs/2203.08792}
}

@misc{fan_hold_2023,
  title = {{HOLD}: Category-agnostic {3D} Reconstruction of Interacting Hands and Objects from Video},
  author = {Fan, Zicong and Parelli, Maria and Kadoglou, Maria Eleni and Kocabas, Muhammed and Chen, Xu and Black, Michael J. and Hilliges, Otmar},
  year = {2023},
  doi = {10.48550/arXiv.2311.18448},
  url = {http://arxiv.org/abs/2311.18448},
  note = {arXiv:2311.18448}
}

@misc{wu_reconstructing_2024,
  title = {Reconstructing Hand-Held Objects in {3D}},
  author = {Wu, Jane and Pavlakos, Georgios and Gkioxari, Georgia and Malik, Jitendra},
  year = {2024},
  url = {http://arxiv.org/abs/2404.06507},
  note = {arXiv:2404.06507}
}

@inproceedings{cotton_biomechanical_2025,
  title = {Biomechanical Reconstruction with Confidence Intervals from Multiview Markerless Motion Capture},
  booktitle = {2025 47th Annual International Conference of the IEEE Engineering in Medicine \& Biology Society (EMBC)},
  author = {Cotton, R. James and Sinz, Fabian},
  year = {2025},
  doi = {10.48550/arXiv.2502.06486},
  url = {http://arxiv.org/abs/2502.06486},
  note = {arXiv:2502.06486}
}

@misc{donahue_embc_2026,
  title = {{EMBC} Special Issue: Calibrated Uncertainty for Trustworthy Clinical Gait Analysis Using Probabilistic Multiview Markerless Motion Capture},
  author = {Donahue, Seth and Djuraskovic, Irina and Shah, Kunal and Sinz, Fabian and Chafetz, Ross and Cotton, R. James},
  year = {2026},
  doi = {10.48550/arXiv.2601.22412},
  url = {http://arxiv.org/abs/2601.22412},
  note = {arXiv:2601.22412}
}
}

\end{document}